\documentclass[10pt,twocolumn,letterpaper]{article}

\usepackage{cvpr}
\usepackage{times}
\usepackage{epsfig}
\usepackage{graphicx}
\usepackage{amsmath}
\usepackage{amssymb}

\usepackage{color}
\usepackage{colortbl}
\usepackage{multirow}
\usepackage{multicol}
\usepackage{arydshln}
\usepackage{adjustbox}

\definecolor{gray}{gray}{0.7}
\definecolor{red}{rgb}{0.9, 0.3, 0.3}
\definecolor{green}{rgb}{0.3, 0.9, 0.3}

\usepackage[pagebackref=true,breaklinks=true,letterpaper=true,colorlinks,bookmarks=false]{hyperref}

\newcommand{\bx}{\mathbf{x}}
\newcommand{\be}{\mathbf{e}}
\newcommand{\by}{\mathbf{y}}
\newcommand{\bw}{\mathbf{w}}
\newcommand{\bW}{\mathbf{W}}
\newcommand{\sign}{\operatornamewithlimits{sign}}

\newcommand\blfootnote[1]{%
  \begingroup
  \renewcommand\thefootnote{}\footnote{#1}%
  \addtocounter{footnote}{-1}%
  \endgroup
}

\cvprfinalcopy 

\def\cvprPaperID{2234} 

\ifcvprfinal\fi
\begin{document}

\title{Learning Visual Features from Large Weakly Supervised Data}

\author{
Armand Joulin$^*$\\
{\tt\small ajoulin@fb.com}
\and
Laurens van der Maaten$^*$\\
{\tt\small lvdmaaten@fb.com}
\and
Allan Jabri\\
{\tt\small ajabri@fb.com}
\and
Nicolas Vasilache\\
{\tt\small ntv@fb.com}
\and
Facebook AI Research\\
770 Broadway, New York NY 10003
}

\begin{multicols}{1}
\maketitle
\end{multicols}

\begin{abstract}
Convolutional networks trained on large supervised dataset produce visual features which form the basis for the state-of-the-art in many computer-vision problems. Further improvements of these visual features will likely require even larger manually labeled data sets, which severely limits the pace at which progress can be made. In this paper, we explore the potential of leveraging massive, weakly-labeled image collections for learning good visual features. We train convolutional networks on a dataset of 100 million Flickr photos and captions, and show that these networks produce features that perform well in a range of vision problems. We also show that the networks appropriately capture word similarity, and learn correspondences between different languages.
\end{abstract}

\section{Introduction}
\blfootnote{* Both authors contributed equally.}
Recent studies have shown that using visual features extracted from convolutional networks trained on large object recognition datasets~\cite{krizhevsky12,simonyan15,szegedy15} can lead to state-of-the-art results on many vision problems including fine-grained classification~\cite{jaderberg15,razavian2014cnn}, object detection~\cite{girshick2014rich}, and segmentation~\cite{pinheiro16}. The success of these networks has been largely fueled by the development of large, manually annotated datasets such as Imagenet~\cite{deng2009imagenet}. This suggests that in order to further improve the quality of visual features, convolutional networks should be trained on larger datasets.

\begin{figure}
\begin{center}
\includegraphics[width=1.05\columnwidth]{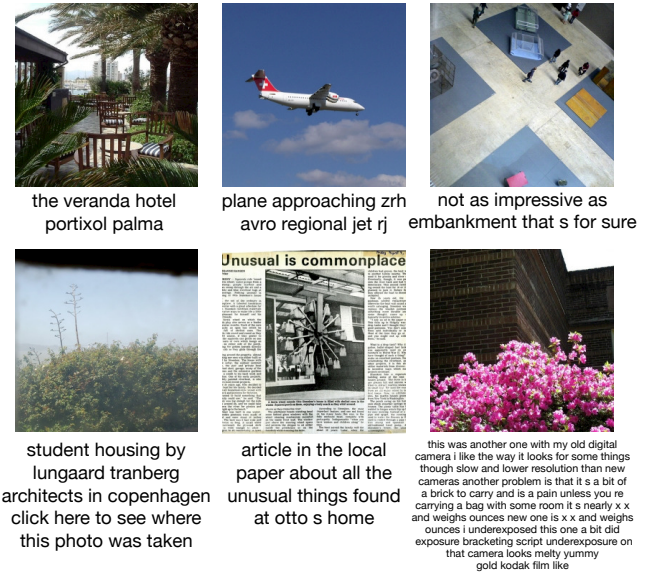}
\end{center}
\vspace{-4mm}
\caption{Six randomly picked photos from the Flickr 100M dataset and the corresponding descriptions.}\label{fig:example_images}
\end{figure}

At the same time, this begs the question whether fully supervised approaches are the right way forward to learning better models. In particular, the manual annotation of ever larger image datasets is very time-consuming\footnote{For instance, the development of the COCO dataset~\cite{lin15} took more than $20,000$ annotator hours spread out over two years.}, which makes it a non-scalable solution to improving recognition performances. Moreover, manually selecting and annotating images often introduces a strong bias towards a specific task~\cite{ponce06,torralba11}. Another problem of fully supervised approaches is that they appear rather inefficient compared to how humans learn to recognize objects: unsupervised and weakly supervised learning plays an important role in human vision~\cite{dicarlo12}, as a result of which humans do not need to see thousands of images of, say, aardvarks to obtain a good grasp of what an aardvark looks like.

In this paper, we depart from the fully supervised learning paradigm and ask the question: \emph{can we learn high-quality visual features from scratch without using any fully supervised data?} 
We perform a series of experiments in which we train models on a large collection of images and their associated captions. This type of data is available in great abundance via photo-sharing websites: specifically, we use a publicly available dataset of 100 million Flickr images and captions \cite{thomee15} (see Figure~\ref{fig:example_images} for six randomly picked Flickr photos and corresponding descriptions). Learning visual representations on such weakly supervised data has three major advantages: (1) there is a near-infinite amount of weakly supervised data available\footnote{The combined number of photo uploads via various platforms was estimated to be 1.8 billion photos per day in 2014 \cite{meeker14}.}, (2) the training data is not biased towards solving a specific task, and (3) it is much more similar to how humans learn to solve vision.

We present experiments showing that convolutional networks can learn to identify words that are relevant to a particular image, despite being trained on very noisy targets. More importantly, our experiments show that the visual features learned by weakly-supervised models are as good as those learned by models that were trained on Imagenet, suggesting that good visual representations can be learned without full supervision. Moreover, our experiments reveal some benefits of training on weakly supervised data such as the Flickr dataset: our models learn word embeddings that are both grounded in vision and capture important semantic information, for instance, on word similarity and analogies. Because our training data is multilingual, our models also relate words from different languages by observing that they are frequently assigned to similar visual inputs. 

\section{Related Work}
\label{Related Work}
This study is not the first to explore alternatives to training convolutional networks on manually annotated datasets~\cite{chen15,divvala2014learning,rubinstein2013unsupervised,zhou2014conceptlearner}. In particular, Chen and Gupta~\cite{chen15} propose a curriculum-learning approach that trains convolutional networks on ``easy'' examples retrieved from Google Images, and then finetunes the models on weakly labeled image-hashtag pairs. Their results suggest that such a two-stage approach outperforms models trained on solely image-hashtag data. This result is most likely due to the limited size of the dataset that was used for training ($\sim$1.2 million images): our results show substantial performance improvements can be obtained by training on much larger image-word datasets. Izadinia \emph{et al.}~\cite{izadinia2015deep} also finetune pretrained convolutional networks on a dataset of Flickr images using a vocabulary of $5,000$ words. By contrast, this study trains convolutional networks \emph{from scratch} on 100 million images associated with $100,000$ words. Ni \emph{et al.}~\cite{ni15} also train convolutional networks on tens of millions of image-word pairs, but their study focuses on systems issues and does not report recognition performances.

Several studies have used weakly supervised data in image-recognition pipelines that use pre-defined visual features. In particular, Li and Fei-Fei~\cite{li10} present a hierarchical topic model that performs simultaneous dataset construction and incremental learning of object recognition models. Li \emph{et al.}~\cite{li13} learn mid-level representations by training a multiple-instance learning variant of SVMs on hand-crafted low-level features extracted from images downloaded using Google Image search. Denton \emph{et al.}~\cite{denton15} learn (user-conditional) embeddings of images and hashtags on a large collection of Instagram photos and hashtags. Torresani \emph{et al.}~\cite{torresani10} train a collection of weak object classifiers and use the classifier outputs as additional image features (in addition to low-level image features). In contrast to these studies, we backpropagate the learning signal through the entire pipeline, allowing us to learn visual features. 

In contrast to our work, many prior studies also attempt to explicitly discard low-quality labels by developing algorithms that identify relevant image-hashtag pairs from a weakly labeled dataset~\cite{fan10,ordonez10,xia14}. These studies solely aim to create a ``clean'' dataset and do not explore the training of recognition pipelines on noisy data. By contrast, we study the training of a full image-recognition pipeline; our results suggest that ``label cleansing'' may not be necessary to learn good visual features if the amount of weakly supervised training data is sufficiently large.

Our work is also related to prior studies 
on multimodal embedding~\cite{socher2013zero,yang2011corpus} that explore approaches such as kernel canonical component analysis~\cite{gong2014multi,hodosh2013framing},
restricted Boltzmann machines~\cite{srivastava2012multimodal},
topic models~\cite{jia2011learning}, and log-bilinear models~\cite{kiros2014multimodal}. 
Some works co-embed images and words~\cite{frome2013devise}, whereas others co-embed images and sentences or n-grams~\cite{farhadi2010every,karpathy2014deep,weston11}.
Frome \emph{et al.}~\cite{frome2013devise} show that convolutional networks trained jointly
on annotated image data and a large corpus of unannotated texts can be
used for zero-shot learning. Our work is different from those prior studies in that we train convolutional networks solely on weakly supervised data.





\section{Weakly Supervised Learning of Convnets}
\label{Weakly Supervised Learning of Convolutional Networks}
We train our models on the publicly available Flickr 100M data set \cite{thomee15}. The data set contains approximately 99.2 million photos with associated titles, hashtags, and captions. We will release our code and models upon publication of the paper.

\noindent\textbf{Preprocessing.} We preprocessed the text by removing all numbers and punctuation (\emph{e.g}., the $\#$ character for hashtags), removing all accents and special characters, and lower-casing. We then used the Penn Treebank tokenizer\footnote{\scriptsize{\url{https://www.cis.upenn.edu/~treebank/tokenizer.sed}}} to tokenize the titles and captions into words, and used all hashtags and words as targets for the photos. We remove the 500 most common words (\emph{e.g.}, ``the'', ``of'', and ``and'') and because the tail of the word distribution is very long \cite{adamic02}, we restrict ourselves to predicting only the $K=\{1,000; 10,000; 100,000\}$ most common words. For these dictionary sizes, the average number of targets per photo is $3.72$, $5.62$, and $6.81$, respectively. The target for each image is a bag of all the words in the dictionary associated with that image, \emph{i.e.}, a multi-label vector $\by \in \{0,1\}^K$. The images were preprocessed by rescaling them to $256\!\times\!256$ pixels, cropping a central region of $224\!\times\!224$ pixels, subtracting the mean pixel value of each image, and dividing each image by the standard deviation of its pixel values.

\noindent\textbf{Network architecture.} We experimented with two convolutional network architectures, \emph{viz.}, the AlexNet architecture \cite{krizhevsky12} and the GoogLeNet architecture \cite{szegedy15}. The AlexNet architecture is a seven-layer architecture that uses max-pooling and rectified linear units at each layer; it has between $15$M and $415$M parameters depending on the vocabulary size. The GoogLeNet architecture is a narrower, twelve-layer architecture that has a shallow auxiliary classifier to help learning; it holds the state-of-the-art on the ImageNet ILSVRC2014 dataset \cite{russakovsky15}. Our GoogLeNet models had between $4$M and $404$M parameters depending on vocabulary size. For exact details on both architectures, we refer the reader to \cite{krizhevsky12} and \cite{szegedy15}, respectively---our architectures only deviate from the architectures described there in the size of their final output layer.

\noindent\textbf{Loss functions.} We denote the training set by $\mathcal{D} = \{ (\bx_n, \by_n) \}_{n = 1,\dots,N}$ with the $D$-dimensional observation $\mathbf{x} \in \mathbb{R}^D$ and the multi-label vector $\by \in \{0,1\}^K$. We parametrize the mapping $f(\bx; \theta)$ from observation $\bx \in \mathbb{R}^D$ to some intermediate embedding $\be \in \mathbb{R}^E$ by a convolutional network with parameters $\theta$; and the mapping from that embedding $\be$ to a label $\by \in \{0,1\}^K$ by $\sign(\bW^\top \be)$, where $\bW$ is an $E \times K$ matrix. The parameters $\theta$ and $\bW$ are optimized jointly to minimize a one-versus-all or multi-class logistic loss. The one-versus-all logistic loss sums binary classifier losses over all classes:

\noindent\resizebox{\columnwidth}{!}{\begin{minipage}{\columnwidth}
\begin{align}
\sum_{n=1}^N  
\sum_{k=1}^K \frac{y_{nk}}{N_k} \log \sigma(f(\bx_n;\theta)) + \frac{1- y_{nk}}{N-N_k} \log (1- \sigma(f(\bx_n,\theta))),\nonumber
\end{align}
\end{minipage}}
\vspace{3mm}  

\noindent where $\sigma(x)= 1 / (1+\exp (-x))$ is the sigmoid function and $N_k$ is the number of positive examples for the class $k$.
The multi-class logistic loss minimizes the negative sum of the log-probabilities over all positive labels. Herein, the probabilities are computed using a softmax layer:

\noindent\resizebox{\columnwidth}{!}{\begin{minipage}{\columnwidth}
\begin{align}
\ell(\theta, \bW; \mathcal{D}) = \frac{-1}{N} \sum_{n=1}^N \sum_{k=1}^K y_{nk} \log \left[ \frac{ \exp(\bw_k^\top f(\bx_n; \theta)) }{ \sum_{k'=1}^K \exp(\bw_{k'}^\top f(\bx_n; \theta)) } \right].\nonumber
\end{align}
\end{minipage}}
\vspace{1mm}  

\noindent In preliminary experiments, we also considered a pairwise ranking loss \cite{usunier2009ranking,weston11}. 
This loss only updates two columns of $\bW$ per training example (corresponding to a positive and a negative label). We found that when training convolutional networks end-to-end, these sparse updates significantly slow down training, which is why we did not consider ranking loss further in this study.

\noindent\textbf{Class balancing.}
The distribution of words in our dataset follows a Zipf distribution~\cite{adamic02}: much of its probability mass is accounted for by a few classes. 
If we are not careful about how we sample training instances, these classes dominate the learning, which may lead to poor general-purpose visual features~\cite{akata14}.
We follow Mikolov \emph{et al.}~\cite{mikolov13} and sample instances \emph{uniformly per class}.
Specifically, we select a training example by picking a word uniformly at random and randomly selecting an image associated with that word. All the other words are considered negative for the corresponding image, \emph{even words that are also associated with that image}. Although this procedure potentially leads to noisier gradients, it works well in practice.

\noindent\textbf{Training.} We trained our models with stochastic gradient descent (SGD) on batches of size $128$. In all experiments, we set the initial learning rate to $0.1$ and after every sweep through a million images (an ``epoch''), we compute the prediction error on a held-out validation set. When the validation error has increased after an ``epoch'', we divide the learning rate by $2$ and continue training; but we use each learning rate for at least $10$ epochs. We stopped training when the learning rate became smaller than $10^{-6}$. AlexNet takes up to two weeks to train on a setup with 4 GPUs, while training a GoogLeNet takes up to three weeks. 

\noindent\textbf{Large dictionary.} Training a network on $100,000$ classes is computationally expensive: a full forward-backward pass through the last linear layer with a single batch takes roughly $1,600$ms (compared to $400$ms for the rest of the network). 
To circumvent this problem, we only update the weights that correspond to classes present in a training batch. This means we update at most $128$ columns of $\bW$ per batch, instead of all $100,000$ columns.
We found such ``stochastic gradient descent over targets'' to work very well in practice: it reduced the training time of our largest models from months to weeks. 

Whilst our stochastic approximation is consistent for the one-versus-all loss, it is not for the multi-class logistic loss:
in the worst-case scenario, the ``approximate'' logistic loss can be arbitrarily far from the true loss. 
However, we observe that the approximation works well in practice, and upper and lower bounds on the expected value of the approximate loss suggest that, indeed, it is closely related to the true loss.
Denoting $s_k = \exp\left(\bw_{k}^\top f(\bx_n; \theta)\right)$ and the set of sampled classes by $\mathcal{C}$ (with $| \mathcal{C} | \leq K$) and leaving out constant terms for brevity, it is trivial to see that the expected approximate loss never overestimates the true loss:
\begin{equation}
\mathbb{E}\left[\log \sum_{c \in \mathcal{C}} s_c \right] \leq \log\left(\sum_{k=1}^K s_k \right) = \log(Z).\nonumber  
\end{equation}

\noindent Assuming that $\forall k:\!s_k\!\geq\!1$\footnote{This assumption can always be satisfied by adding a constant inside the exponentials of both the numerator and the denominator of the softmax.}, we use Markov's inequality to obtain a lower bound on the expected approximate loss, too:
\noindent\resizebox{\columnwidth}{!}{\begin{minipage}{\columnwidth}
\begin{align}
\mathbb{E}\left[\log \sum_{c \in \mathcal{C}} s_c\right] \geq P\left(\frac{1}{|  \mathcal{C} |} \sum_{c \in \mathcal{C}} s_c \geq \frac{1}{K} Z \right)\left(\log \frac{|  \mathcal{C} |}{K} + \log Z \right)\nonumber.
\end{align}
\end{minipage}}
\vspace{2mm}

\noindent This bound relates the sample average of $s_c$ to its expected value, and is exact when $|\mathcal{C}|\!\rightarrow\!K$. The lower bound only contains an additive constant $\log(|\mathcal{C}| / K)$, which shows that the approximate loss is closely related to the true loss.

\begin{table}
\begin{center}
\begin{small}
\setlength\tabcolsep{2.5pt}
\begin{tabular}{|l|l||c|c|c|}
\hline
\multicolumn{2}{|c||}{} & \multicolumn{3}{c|}{\textbf{Dictionary size K}}\\
\textbf{Type} & \textbf{Network} & $\mathbf{1,000}$ & $\mathbf{10,000}$ & $\mathbf{100,000}$ \\
\hline\hline
\rowcolor{gray}\cellcolor{white} & AlexNet & 8.27 & 4.01 & 1.61 \\
\multirow{-2}{*}{\cellcolor{white} Pretrained} & GoogLeNet & 13.20 & 4.76 & 1.54 \\\hdashline
\rowcolor{gray}\cellcolor{white} & AlexNet & 17.98 & 6.27 & 2.56  \\
\multirow{-2}{*}{\cellcolor{white} End-to-end} & GoogLeNet & 20.21 & 6.47 & -- \\ 
\hline
\end{tabular}
\end{small}
\end{center}
\caption{Precision@10 on held-out test data of word prediction models on the Flickr 100M dataset for three different dictionary sizes $K$. We present results for (1) logistic regressors trained on features extracted from convolutional networks that were \emph{pretrained} on Imagenet and (2) convolutional networks trained \emph{end-to-end} using multiclass logistic loss. Higher values are better.}
\label{table:hp_results}
\end{table}

\section{Experiments}
\label{Experiments}
To assess the quality of our weakly-supervised convolutional networks, we performed three sets of experiments: (1) experiments measuring the ability of the models to predict words given an image, (2) transfer-learning experiments measuring the quality of the visual features learned by our models in a range of computer-vision tasks, and (3) experiments evaluating the quality of the word embeddings learned by the networks.

\begin{figure}
\begin{center}
\includegraphics[width=\columnwidth]{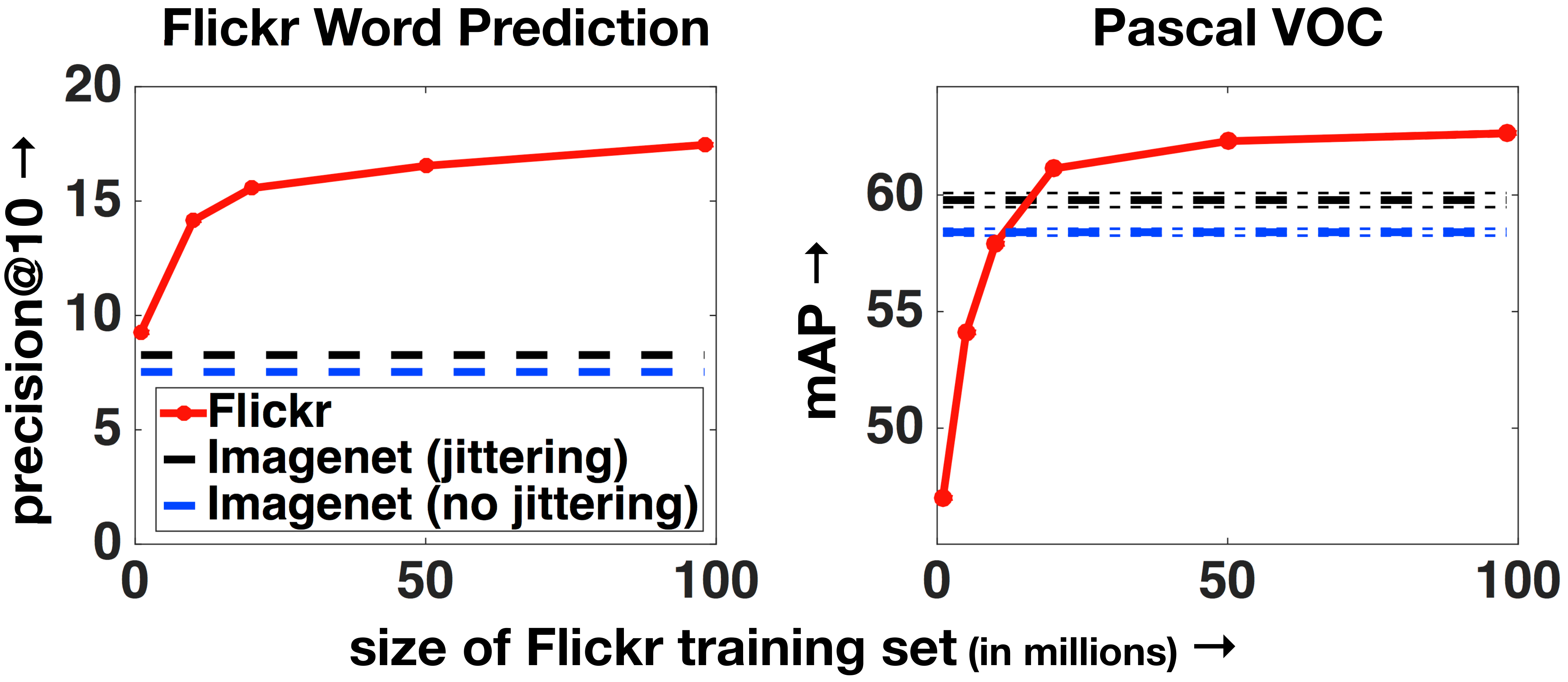}
\end{center}
\caption{\emph{Lefthand side:} Precision@10 of by weakly supervised AlexNets trained on Flickr datasets of different sizes on a held-out test set, using $K\!=\!1,000$ (in red) and a single crop. For reference, we also show the precision@10 of logistic regression trained on features from convolutional networks trained on ImageNet with and without jittering (in blue and black, respectively). \emph{Righthand side:} Mean average precision on Pascal VOC 2007 dataset obtained by logistic regressors trained on features extracted from AlexNet trained on Flickr (in red) and ImageNet with and without jittering (in blue and black). Higher values are better.}\label{fig:hp_learning_curve}
\end{figure}

\begin{figure}
\includegraphics[width=\columnwidth]{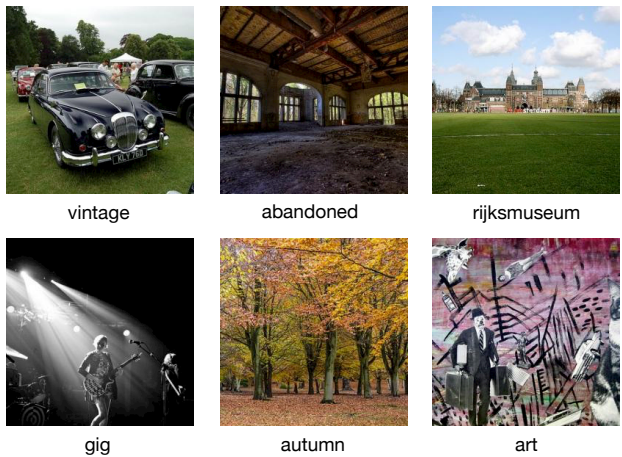}
\caption{Six test images with high scores for different words. The scores were computed using an AlexNet trained on the Flickr dataset with a dictionary size of $K\!=\!100,000$.}\label{fig:example_predictions}
\end{figure}

\subsection{Experiment 1: Associated Word Prediction}
\label{Experiment 1}

\noindent\textbf{Experimental setup.} We measure the ability of our models to predict words that are associated with an image using the precision@k on a test set of 1 million Flickr images, which we held out until after all our models were trained. Precision@k is a suitable measure for assessing word prediction performance because (1) it corresponds naturally to use cases in which a user retrieves images using a text query and  inspects only the top k results and (2) it is robust to the fact that targets are noisy, \emph{i.e.}, that images may have words assigned to them that do not describe their visual content.

\begin{figure*}
\includegraphics[width=.95\textwidth]{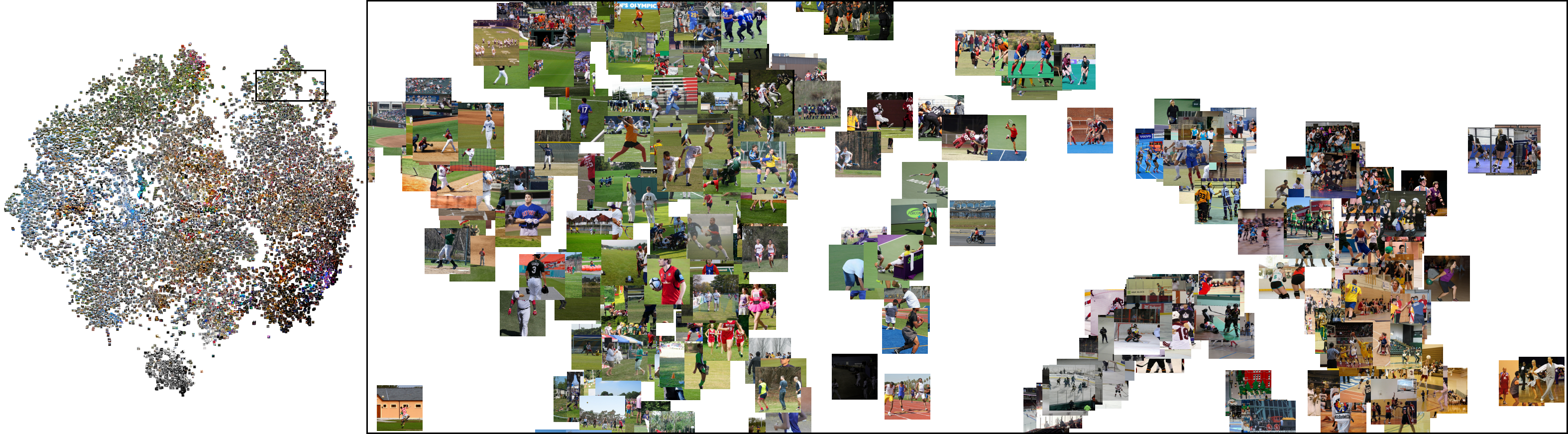}
\caption{t-SNE map of $20,000$ Flickr test images based on features extracted from the last layer of an AlexNet trained with $K\!=\!1,000$. A full-resolution map is presented in the supplemental material. The inset shows a cluster of sports.}\label{fig:tsne_images}
\end{figure*}

\noindent\textbf{Results.} Table~\ref{table:hp_results} presents the precision@10 of word prediction models trained on the Flickr dataset using dictionaries with $1,000$, $10,000$, and $100,000$ words\footnote{Our GoogLeNet networks with $K\!=\!100,000$ words did not finish training by the submission deadline. We will update the paper with those results as they become available.}. As a baseline, we train L2-regularized logistic regressors on features produced by convolutional networks trained on the Imagenet dataset\footnote{The Imagenet models were trained $224\!\times\!224$ crops that where randomly selected from $256\!\times\!256$ pixel input images. We applied photometric jittering on the input images \cite{howard13}, and trained using SGD with batches of $128$ images. Our pretrained networks perform on par with the state-of-the-art on ImageNet: a single AlexNet obtains a top-5 test error of $24.0\%$ on a single crop, and our GoogLeNet obtains a top-5 error of $10.7\%$.}; the regularization parameter was tuned on a held-out validation set.
The results of this experiment show that end-to-end training of convolutional networks on the Flickr dataset works substantially better than training a classifier on features extracted from an Imagenet-pretrained network: end-to-end training leads to a relative gain of $45$ to $110$\% in precision@10. This suggests that the features learned by networks on the Imagenet dataset are too tailored to the specific set of classes in that dataset. The results also show that the relative differences between the GoogLeNet and AlexNet architectures are smaller on the Flickr 100M dataset than on the Imagenet dataset, possibly, because GoogLeNet has less capacity than AlexNet.

In preliminary experiments, we also trained models using one-versus-all logistic loss: using a dictionary of $K\!=\!1,000$ words, such a model achieves a precision@10 of $16.43$ (compared to $17.98$ for multiclass logistic loss). We surmise this is due to the problems one-versus-all logistic loss has in dealing with class imbalance: because the number of negative examples is much higher than the number of positive examples (for the most frequent class, more than $95.0\%$ of the data is still negative), the rebalancing weight in front of the positive term is very high, which leads to spikes in the gradient magnitude that hamper SGD training. We tried various reweighting schemes to counter this effect, but nevertheless, multiclass logistic loss consistently outperformed one-versus-all logistic loss in our experiments.

To investigate the performance of our models as a function of the amount of training data, we also performed experiments in which we varied the Flickr training set size. The lefthand side of Figure~\ref{fig:hp_learning_curve} presents the resulting learning curves for the AlexNet architecture with $K\!=\!1,000$. The figure shows that there is a clear benefit of training on larger datasets: the word prediction performance of the networks increases substantially when the training set is increased beyond 1 million images (which is roughly the size of Imagenet); for our networks, it only levels out after $\sim\!50$ million images. 

To illustrate the kinds of words for which our models learn good representations, we show a high-scoring test image for six different words in Figure~\ref{fig:example_predictions}. To obtain more insight into the features learned by the models, we applied t-SNE \cite{vandermaaten14,vandermaaten08} to features extracted from the penultimate layer of an AlexNet trained on $1,000$ words. This produces maps in which images with similar visual features are close together; Figure~\ref{fig:tsne_images} shows such a map of $20,000$ Flickr test images. The inset shows a ``sports'' cluster that was formed by the visual features; interestingly, it contains visually very dissimilar sports ranging from baseball to field hockey, ice hockey and rollerskating. Whilst all sports are grouped together, the individual sports are still clearly separable: the model can capture this multi-level structure because the images sometimes occur with the word ``sports'' and sometimes with the name of the individual sport itself. A model trained on classification datasets such as Pascal VOC is unlikely to learn similar structure unless an explicit target taxonomy is defined (as in the Imagenet dataset). Our results suggest that such taxonomies can be learned from weakly labeled data instead.

\subsection{Experiment 2: Transfer Learning}
\label{Experiment 2}

\begin{table*}
\begin{center}
\setlength\tabcolsep{2.5pt}
\begin{adjustbox}{max width=\textwidth}
\begin{tabular}{|l||l||c|c|c|c|c|c|c|c|c|c|c|c|c|c|c|c|c|c|c|c||c|}
\hline
\textbf{Dataset} & \textbf{Model} & 
\raisebox{-.5\height}{\includegraphics[width=0.05\textwidth]{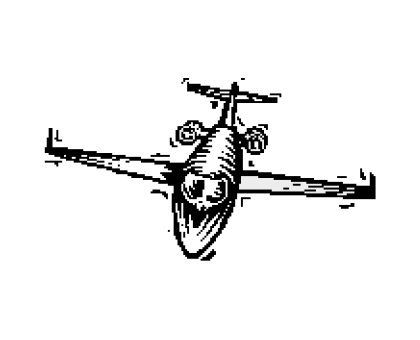}} & 
\raisebox{-.5\height}{\includegraphics[width=0.05\textwidth]{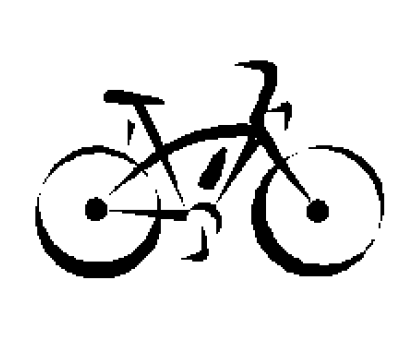}} & 
\raisebox{-.5\height}{\includegraphics[width=0.05\textwidth]{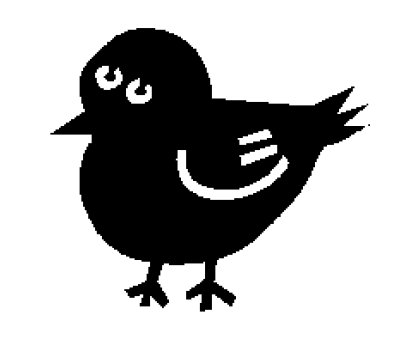}} & 
\raisebox{-.5\height}{\includegraphics[width=0.05\textwidth]{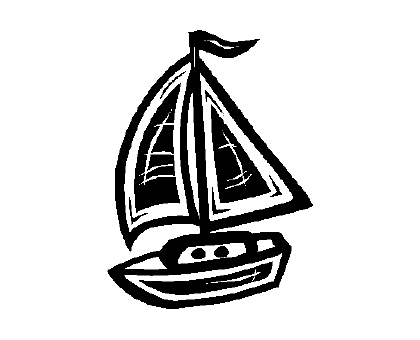}} & 
\raisebox{-.5\height}{\includegraphics[width=0.05\textwidth]{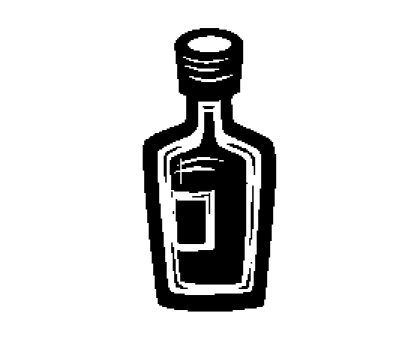}} & 
\raisebox{-.5\height}{\includegraphics[width=0.05\textwidth]{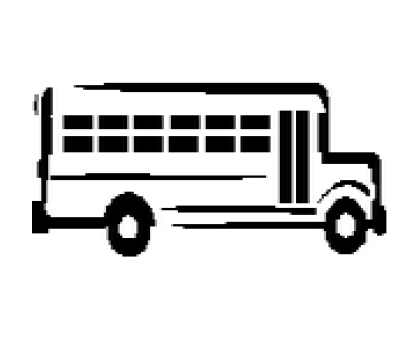}} & 
\raisebox{-.5\height}{\includegraphics[width=0.05\textwidth]{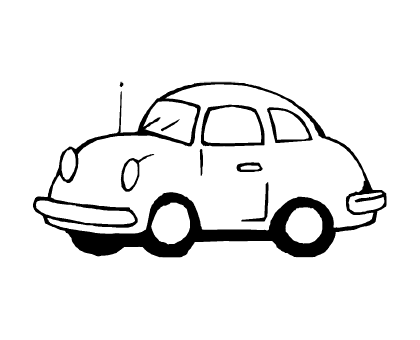}} & 
\raisebox{-.5\height}{\includegraphics[width=0.05\textwidth]{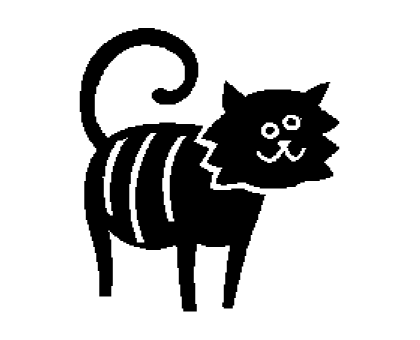}} &
\raisebox{-.5\height}{\includegraphics[width=0.05\textwidth]{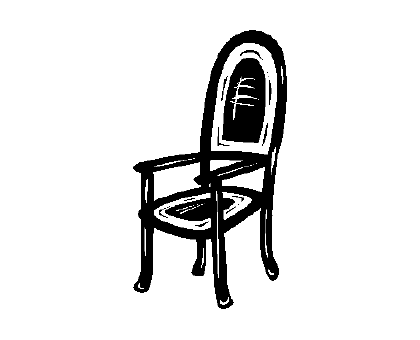}} & 
\raisebox{-.5\height}{\includegraphics[width=0.05\textwidth]{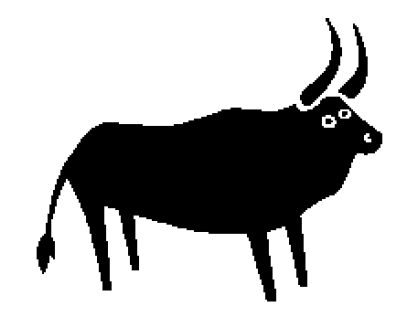}} &
\raisebox{-.5\height}{\includegraphics[width=0.05\textwidth]{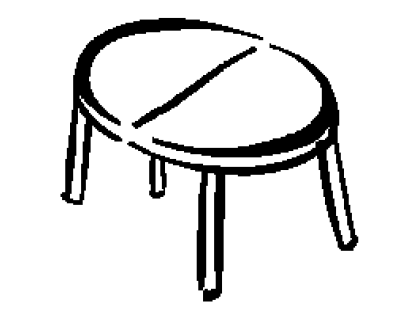}} & 
\raisebox{-.5\height}{\includegraphics[width=0.05\textwidth]{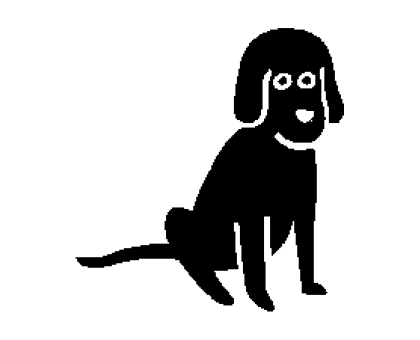}} & 
\raisebox{-.5\height}{\includegraphics[width=0.05\textwidth]{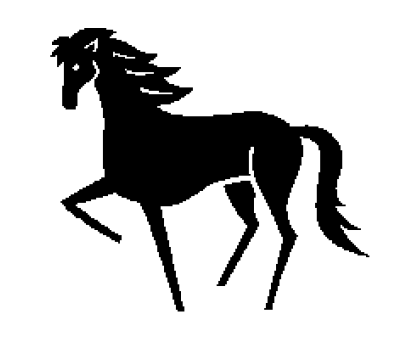}} & 
\raisebox{-.5\height}{\includegraphics[width=0.05\textwidth]{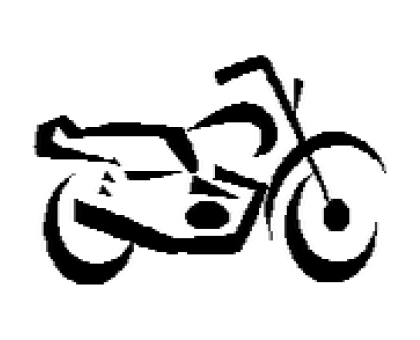}} & 
\raisebox{-.5\height}{\includegraphics[width=0.05\textwidth]{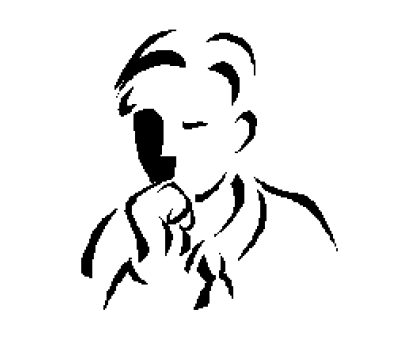}} & 
\raisebox{-.5\height}{\includegraphics[width=0.05\textwidth]{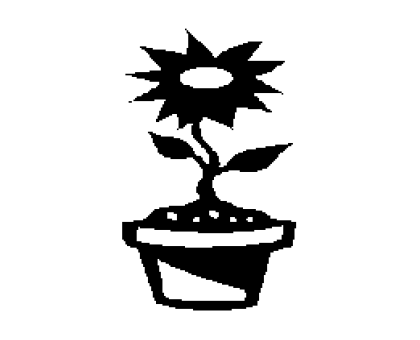}}
& \raisebox{-.5\height}{\includegraphics[width=0.05\textwidth]{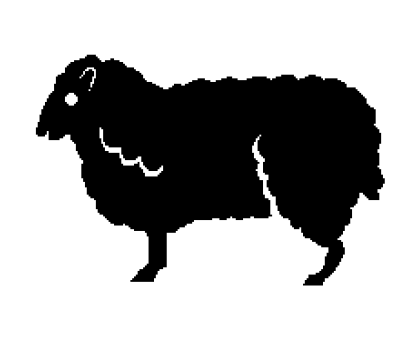}} & 
\raisebox{-.5\height}{\includegraphics[width=0.05\textwidth]{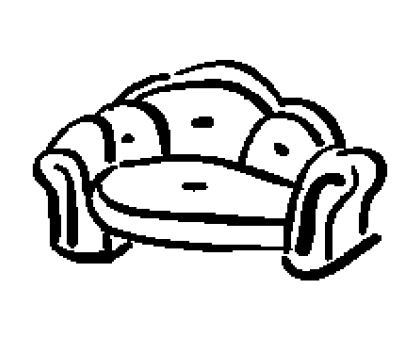}} & 
\raisebox{-.5\height}{\includegraphics[width=0.05\textwidth]{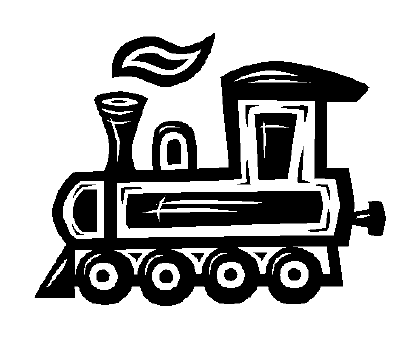}} & 
\raisebox{-.5\height}{\includegraphics[width=0.05\textwidth]{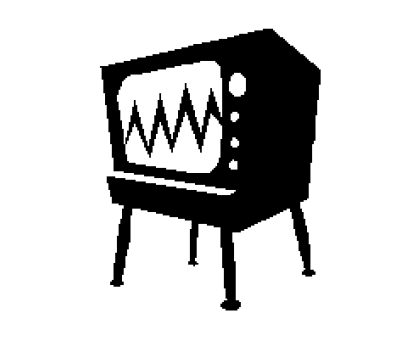}} & \textbf{mAP} \\
\hline\hline
\rowcolor{gray}\cellcolor{white} & AlexNet & 75.7 & 61.9 & 66.9 & 66.5 & 29.3 & 56.1 & 73.5 & 68.0 & 47.1 & 40.9 & 57.4 & 60.0 & 74.0 & 63.2 & 86.2 & 38.8 & 57.9 & 45.5 & 75.7 & 51.1 & 59.8 \\
\multirow{-2}{*}{Imagenet} & GoogLeNet & 91.3 & 84.0 & 88.4 & 87.2 & 42.4 & 79.6 & 87.3 & 85.0 & 59.1 & 66.5 & 69.5 & 83.3 & 86.6 & 82.9 & 88.4 & 57.5 & 75.8 & 64.6 & 89.5 & 73.8 & 77.1  \\\hdashline
\rowcolor{gray}\cellcolor{white} & AlexNet & 84.0 & 72.2 & 70.2 & 77.0 & 29.5 & 60.8 & 79.3 & 69.5 & 49.2 & 40.5 & 54.0 & 57.1 & 79.2 & 64.6 & 90.2 & 43.0 & 47.5 & 44.1 & 85.0 & 50.7 & 62.4 \\
\multirow{-2}{*}{Flickr} & GoogLeNet & 91.5 & 83.7 & 84.1 & 88.5 & 41.7 & 78.0 & 86.8 & 84.0 & 54.7 & 55.5 & 63.3 & 78.5. & 86.0 & 77.4 & 91.1 & 51.3 & 60.8 & 52.7 & 91.9 & 60.9 & 73.2 \\\hdashline
\rowcolor{gray}\cellcolor{white} & AlexNet & 82.96 & 70.32 & 73.28 & 76.29 & 32.21 & 61.84 & 79.81 & 72.91 & 51.56 & 43.82 & 60.77 & 63.32 & 78.63 & 67.72 & 90.26 & 45.45 & 53.15 & 49.14 & 84.8 & 55.8 & 64.7 \\
\multirow{-2}{*}{Combined} & GoogLeNet & 94.09 & 85.03 & 89.71 & 88.47 & 49.35 & 81.47 & 88.1 & 85.2 & 60.51 & 68.37 & 71.65 & 85.81 & 88.87 & 85.22 & 88.69 & 60.45 & 77.26 & 66.61 & 90.71 & 74.49 & 79.0 \\  
\hline
\end{tabular}
\end{adjustbox}
\end{center}
\caption{Pascal VOC 2007 dataset: Average precision (AP) per class and mean average precision (mAP) of classifiers trained on features extracted with networks trained on the Imagenet and the Flickr dataset (using $K\!=\!1,000$ words). Higher values are better.}
\label{table:transfer_VOC07}
\end{table*}

\noindent\textbf{Experimental setup.} To assess the quality of the visual features learned by our models, we performed transfer-learning experiments on seven test datasets comprising a range of computer-vision tasks: (1) the MIT Indoor dataset \cite{quattoni09}, (2) the MIT SUN dataset \cite{xiao10}, (3) the Stanford 40 Actions dataset \cite{yao11}, (4) the Oxford Flowers dataset \cite{nilsback08}, (5) the Sports dataset \cite{gupta09}, (6) the ImageNet ILSVRC 2014 dataset \cite{russakovsky15}, and (7) the Pascal VOC 2007 dataset \cite{everingham15}. We applied the same preprocessing as before on all datasets: we resized the images to $224\!\times\!224$ pixels, subtracted their mean pixel value, and divided by their standard deviation.

Following \cite{razavian14}, we compute the output of the penultimate layer for an input image and use this output as a feature representation for the corresponding image. We evaluate features obtained from Flickr-trained networks as well as Imagenet-trained networks, and we also perform experiments where we combine both features by concatenating them. We train L2-regularized logistic regressors on the features to predict the classes corresponding to each of the datasets. For all datasets except the Imagenet and Pascal VOC datasets, we report classification accuracies on a separate, held-out test set. For Imagenet, we report classification errors on the validation set. For Pascal VOC, we report average precisions on the test set as is customary for that dataset. As before, we use convolutional networks trained on the Imagenet dataset as baseline. Additional details on the setup of the transfer-learning experiments are presented in the supplemental material.

\begin{figure}
\includegraphics[width=1.05\columnwidth]{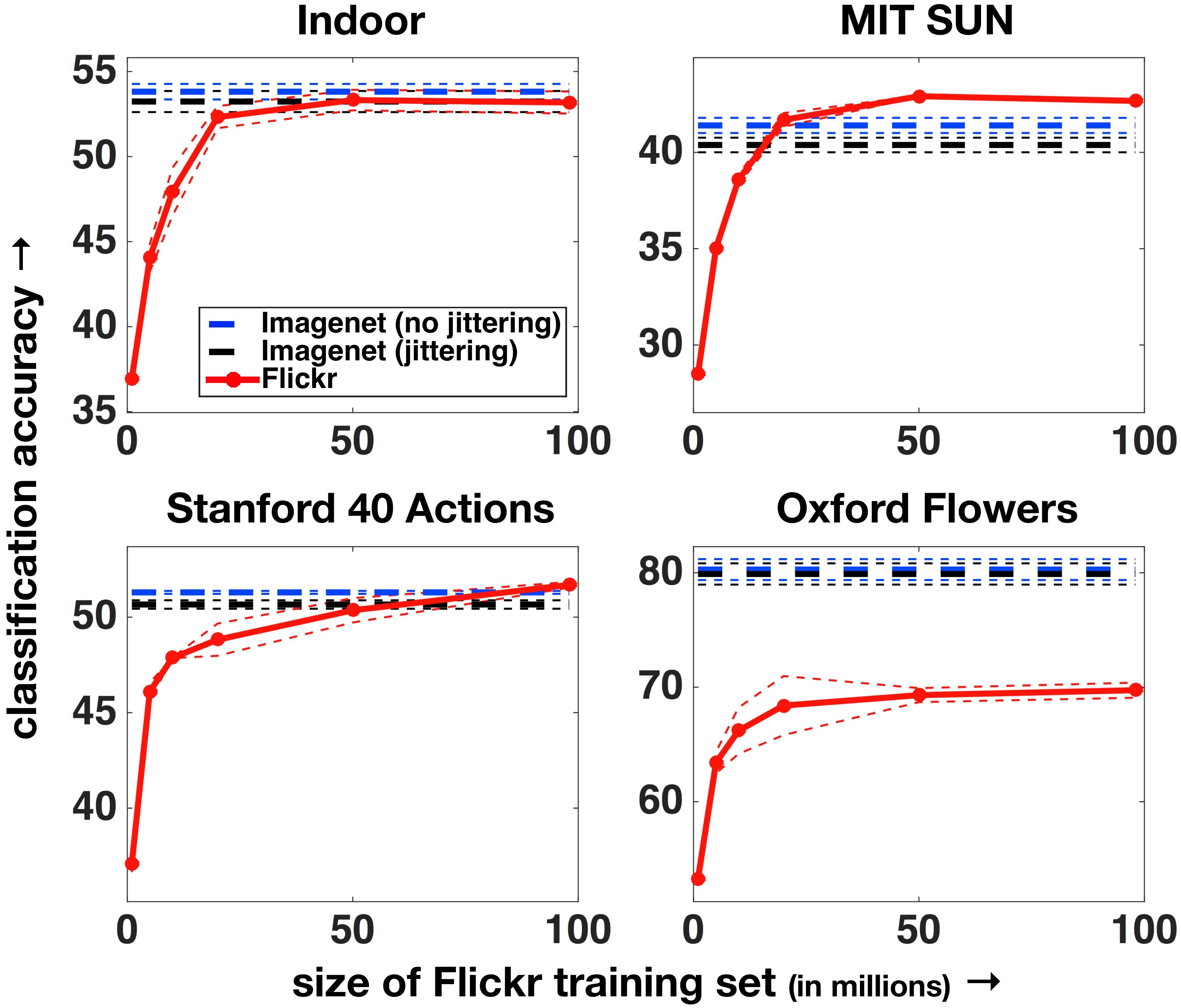}
\caption{Average classification accuracy (averaged over ten runs) of logistic regressors trained on features produced by weakly supervised AlexNets trained on Flickr image-caption datasets of different sizes on six different datasets (in red). For reference, we also show the classification accuracy of classifiers trained on features from convolutional networks trained on ImageNet without jittering (in black) and with jittering (in blue). Dashed lines indicate the standard deviation across runs. Higher values are better.}\label{fig:transfer_learning_curve}
\end{figure}

\noindent\textbf{Results.} Table~\ref{table:transfer_results} presents the classification accuracies---averaged over $10$ runs---of logistic regressors on six datasets for both fully supervised and weakly supervised feature-production networks, as well as for a combination of both networks. Table~\ref{table:transfer_VOC07} presents the average precision on the Pascal VOC 2007 dataset. Our weakly supervised models were trained on a dictionary of $K\!=\!1,000$ words (we obtained similar results for models trained on $10,000$ and $100,000$ words; see the supplementary material). The results in the tables show that using the AlexNet architecture, weakly supervised networks learn visual features of similar quality as fully supervised networks. This is quite remarkable because the networks learned these features \emph{without any strong supervision}. 

Admittedly, weakly supervised networks perform poorly on the flowers dataset: Imagenet-trained networks produce better features for that dataset, presumably, because the Imagenet dataset itself focuses strongly on fine-grained classification. Interestingly, fully supervised networks do learn better features than weakly supervised networks when a GoogLeNet architecture is used: this result is in line with the results from~\ref{Experiment 1}, which suggest that GoogLeNet has too little capacity to learn optimal models on the Flickr data. The substantial performance improvements we observe in experiments in which features from both networks are combined suggest that the features learned by both models complement each other. We note that achieving state-of-the-art results~\cite{chatfield2011devil,oquab2014learning,razavian2014cnn,zhou2014learning} on these datasets requires the development of tailored pipelines, \emph{e.g.}, using many image transformations and model ensembles, which is out of the scope of this paper.

\begin{table}
\begin{center}
\begin{small}
\setlength\tabcolsep{2.5pt}
\begin{adjustbox}{max width=\columnwidth}
\begin{tabular}{|l|l||c|c|c|c|c|c|c|c|c|}
\hline
\textbf{Dataset} & \textbf{Model} & \textbf{Indoor} & \textbf{SUN} & \textbf{Action} & \textbf{Flower} & \textbf{Sports} & \textbf{ImNet}\\
\hline\hline
\rowcolor{gray}\cellcolor{white} & AlexNet & 53.82 & 41.40 & 51.27 & 80.28 & 86.07 & 53.63\\
\multirow{-2}{*}{Imagenet} & GoogLeNet & 64.00 & 48.76 & 67.10 & 79.05 & 95.91 & 69.89\\\hdashline
\rowcolor{gray}\cellcolor{white} & AlexNet & 53.19 & 42.67 & 51.69 & 69.72 & 86.79 & 34.93 \\
\multirow{-2}{*}{Flickr} & GoogLeNet & 55.56 & 44.43 & 52.84 & 65.80 & 87.40 & 33.62 \\\hdashline
\rowcolor{gray}\cellcolor{white} & AlexNet & 58.76 & 47.27 & 56.35 & 83.28 & 87.50 & -- \\
\multirow{-2}{*}{Combined} & GoogLeNet & 67.87 & 55.04 & 69.19 & 83.74 & 95.79 & -- \\\hline
\end{tabular}
\end{adjustbox}
\end{small}
\end{center}
\caption{Classification accuracies on held-out test data of L2-regularized logistic regressors obtained on six datasets (MIT Indoor, MIT SUN, Stanford 40 Actions, Oxford Flowers, Sports, and ImageNet) based on feature representations obtained from convolutional networks trained on the Imagenet and the Flickr dataset (using $K=1,000$ words and a single crop). Errors are averaged over $10$ runs. Higher values are better.} 
\label{table:transfer_results}
\end{table}

We also measured the transfer-learning performance as a function of the Flickr training set size. The results of these experiments with the AlexNet architecture and $K\!=\!1,000$ are presented in Figure~\ref{fig:transfer_learning_curve} for four of the datasets (Indoor, MIT SUN, Stanford 40 Actions, and Oxford Flowers); and in the righthand side of Figure~\ref{fig:hp_learning_curve} for the Pascal VOC dataset. The results are in line with those in~\ref{Experiment 1}: they show that tens of millions of images are required to learn good feature-production networks on weakly supervised data.

\subsection{Experiment 3: Assessing Word Embeddings}
\label{Experiment 3}
The weights in the last layer of our networks can be viewed as an embedding of the words. This word embedding is, however, different from those learned by language models such as word2vec \cite{mikolov13} that learn embeddings based on word co-occurrence: it is constructed without ever observing two words co-occurring (recall that during training, we use a single, randomly selected word as target for an image). This means that structure in the word embedding can only be learned when the network notices that two words are assigned to images with a similar visual structure. We perform two sets of experiments to assess the quality of the word embeddings learned by our networks: (1) experiments investigating how well the word embeddings represent semantic information and (2) experiments investigating the ability of the embeddings to learn correspondences between different languages.

\noindent\textbf{Semantic information.} We evaluate our word embeddings on two datasets that capture different types of semantic information: (1) a syntactic-semantic questions dataset \cite{mikolov13} and (2) the MEN word similarity dataset \cite{bruni12}. The syntactic-semantic dataset contains $8,869$ semantic and $10,675$ syntactic questions of the form ``A is to B as C is to D''. Following \cite{mikolov13}, we predict D by finding the word embedding vector $\bw_D$ that has the highest cosine similarity with $\bw_B\!-\!\bw_A\!+\!\bw_C$ (excluding A, B, and C from the search), and measure the number of times we predict the correct word D. The MEN dataset contains $3,000$ word pairs spanning $751$ unique words---all of which appear in the ESP Game image dataset---with an associated similarity rating. The similarity ratings are averages of ratings provided by a dozen human annotators. Following \cite{kiela14} and others, we measure the quality of word embeddings by the Spearman's rank correlation of the cosine similarity of the word pairs and the human-provided similarity rating for those pairs. In all experiments, we excluded word quadruples / pairs that contained words that are not in our dictionary. We repeated the experiments for three dictionary sizes. As a baseline, we measured the performance of word2vec models that were trained on all comments in the Flickr dataset (using only the words in the dictionary).

The prediction accuracies of our experiments on the syntactic-semantic dataset for three dictionary sizes are presented in Table~\ref{table:w2v_results}. Table~\ref{table:wordsim_results} presents the rank correlations for our word embeddings on the MEN dataset (for three vocabulary sizes). As before, we only included word pairs for which both words appeared in the vocabulary. The results of these experiments show that our weakly supervised models, indeed, learned meaningful semantic structure. The results also show that the quality of our word embeddings is lower than that of word2vec, because unlike our models, word2vec observes word co-occurrences during training.

\begin{table}
\begin{center}
\begin{small}
\setlength\tabcolsep{2.5pt}
\begin{adjustbox}{max width=\columnwidth}
\begin{tabular}{|l||c|c|c|}
\hline
\textbf{Model} & $\mathbf{K\!=\!1,000}$ & $\mathbf{K\!=\!10,000}$ & $\mathbf{K\!=\!100,000}$\\
\hline\hline
\rowcolor{gray} AlexNet & 67.91 & 29.29 & 0.85\\
GoogLeNet & 71.92 & 24.06 & -- \\
\rowcolor{gray} word2vec  & 71.92 & 61.35 & 47.24\\\hdashline
AlexNet + word2vec & 74.79 & 57.26 & 44.35\\
\rowcolor{gray} GoogLeNet + word2vec & 75.36 & 56.05 & -- \\
\hline
\end{tabular}
\end{adjustbox}
\end{small}
\end{center}
\caption{Prediction accuracy of predicting D in questions ``A is to B like C is to D'' using convolutional-network word embeddings and word2vec on the syntactic-semantic dataset, using three dictionary sizes. Questions containing words not in the dictionary were removed. Higher values are better.}
\label{table:w2v_results}
\end{table}

\begin{table}
\begin{center}
\begin{small}
\setlength\tabcolsep{2.5pt}
\begin{adjustbox}{max width=\columnwidth}
\begin{tabular}{|l||c|c|c|}
\hline
\textbf{Model} & $\mathbf{K\!=\!1,000}$ & $\mathbf{K\!=\!10,000}$ & $\mathbf{K\!=\!100,000}$\\
\hline\hline
\rowcolor{gray}AlexNet & 73.77 & 75.73 & 67.35\\
GoogLeNet & 75.72 & 75.89 & -- \\
\rowcolor{gray}word2vec  & 75.25 & 77.53 & 77.91\\\hdashline
 AlexNet + word2vec & 78.17 & 79.24 & 78.57 \\
\rowcolor{gray}GoogLeNet + word2vec & 78.75 & 79.11 & -- \\
\hline
\end{tabular}
\end{adjustbox}
\end{small}
\end{center}
\caption{Spearman's rank correlation of cosine similarities between convolutional-network (and word2vec) word embeddings and human similarity judgements on the MEN dataset. Word pairs containing words not in the dictionary were removed. Higher values are better.}
\label{table:wordsim_results}
\end{table}

\begin{figure*}
\includegraphics[width=0.95\textwidth]{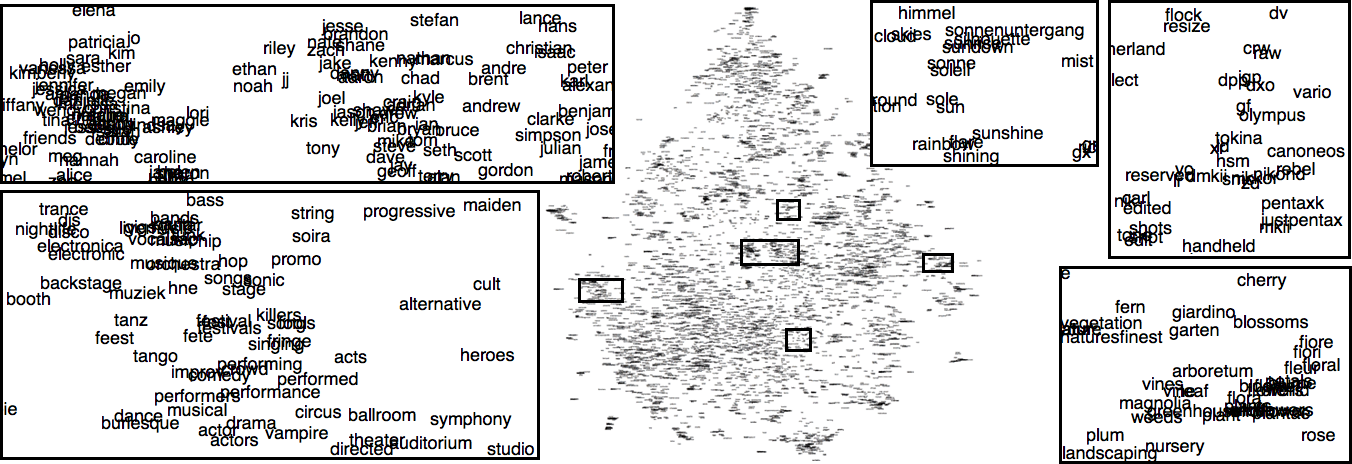}
\caption{t-SNE map of $10,000$ words based on their embeddings as learned by a weakly supervised convolutional network trained on the Flickr dataset. Note that all the semantic information represented in the word embeddings is the result of observing that these words are assigned to images with similar visual content (the model did not observe word co-occurrences during training). A full-resolution version of the map is provided in the supplemental material.}\label{fig:hashtag_tsne}
\end{figure*}

We also made t-SNE maps of the embedding of $10,000$ words in Figure~\ref{fig:hashtag_tsne}. The insets highlight six ``topics'': (1) musical performance, (2) sunsets, (3) female and male first names, (4) gardening, (5) photography, and (6) military. These topics were identified solely because the words in them are associated with images containing similar visual content: for instance, first names are likely to be assigned to photos showing one or a few persons. Interestingly, the ``sunset'' and ``gardening'' topics show examples of grouping of words from different languages. For instance, ``sonne'', ``soleil'', ``sole'' mean ``sun'' in German, French, and Italian, respectively; and ``garten'' and ``giardino'' are the German and Italian words for garden.

\noindent\textbf{Multi-lingual correspondences.} To further investigate the ability of our models to find correspondences between words from different languages, we selected pairs of words from an English-French dictionary\footnote{\scriptsize{\url{http://www-lium.univ-lemans.fr/~schwenk/nnmt-shared-task/}}} for which: (1) both the English and the French word are in the Flickr dictionary and (2) the English and the French word are different. This produced $309$ English-French word pairs for models trained on $K\!=\!10,000$ words, and $3,008$ English-French word pairs for models trained on $K\!=\!100,000$ words. We measured the quality of the multi-lingual word correspondences in the embeddings by taking a word in one language and ranking the words in the other language according to their cosine similarity with the query word. We measure the precision@k of the predicted word ranking, using both English and French words as query words.

\begin{table}
\begin{center}
\begin{small}
\setlength\tabcolsep{2.5pt}
\begin{tabular}{|l|l||c|c|c|}
\hline
\textbf{K} & \textbf{Query} $\rightarrow$ \textbf{Response} & $\mathbf{k=1}$ & $\mathbf{k=5}$ & $\mathbf{k=10}$\\
\hline\hline
\rowcolor{gray} \cellcolor{white} & English $\rightarrow$ French & 33.01 & 50.16 & 55.34\\
\multirow{-2}{*}{$10,000$} & French $\rightarrow$ English & 23.95 & 50.16 & 56.63\\\hdashline
\rowcolor{gray}\cellcolor{white} & English $\rightarrow$ French & 12.30 & 22.24 & 26.50\\
\multirow{-2}{*}{$100,000$} & French $\rightarrow$ English & 10.11 & 18.78 & 23.44\\ 
\hline
\end{tabular}
\end{small}
\end{center}
\caption{Precision@k of identifying the French counterpart of an English word (and vice-versa) for two dictionary sizes, at three different levels of k. Chance level (with $k\!=\!1$) is $0.0032$ for $K\!=\!10,000$ words and $0.00033$ for $K\!=\!100,000$ words. Higher values are better.}
\label{table:translation_results}
\end{table}

Table~\ref{table:translation_results} presents the results of this experiment: for a non-trivial number of words, our procedure correctly identified the French translation of an English word, and vice versa. Finding the English counterpart of a French word is harder then the other way around, presumably, because there are more English than French words in the dictionary: this means the English word embeddings are better optimized than the French ones. In Table~\ref{table:translation_examples}, we show the ten most similar word pairs, measured by the cosine similarity between their word embeddings. These word pairs suggest that models trained on Flickr data find correspondences between words that have clear visual representations, such as ``tomatoes'' or ``bookshop''. Interestingly, the identified English-French matches appear to span a broad set of domains, including objects such as ``pencils'', locations such as ``mauritania'', and concepts such as ``infrared''.

\begin{table}
\begin{center}
\begin{small}
\setlength\tabcolsep{2.5pt}
\begin{tabular}{|c|c||c|c|}
\hline
\textbf{English} & \textbf{French} & \textbf{English} & \textbf{French}\\
\hline\hline
\cellcolor{green} oas      & \cellcolor{green} oea          & \cellcolor{green} uzbekistan & \cellcolor{green} ouzbekistan\\
\cellcolor{green} infrared & \cellcolor{green} infrarouge   & \cellcolor{green} mushroom & \cellcolor{green} champignons  \\
\cellcolor{green} tomatoes & \cellcolor{green} tomates      & \cellcolor{red} filmed  & \cellcolor{red} serveur           \\
\cellcolor{green} bookshop & \cellcolor{green} librairie    & \cellcolor{green} mauritania  & \cellcolor{green} mauritanie\\
\cellcolor{red} server     & \cellcolor{red} apocalyptique  & \cellcolor{green} pencils   & \cellcolor{green} crayons     \\
\hline
\end{tabular}
\end{small}
\end{center}
\caption{Ten highest-scoring pairs of words, as measured by the cosine similarity between the corresponding word embeddings. Correct pairs of words are colored green, and incorrect pairs are colored red according to the dictionary. The word ``oas'' is an abbreviation for the Organization of American States.}
\label{table:translation_examples}
\end{table}


\section{Discussion and Future Work}
\label{Discussion}
This study demonstrates that convolutional networks can be trained \emph{from scratch} without any manual annotation and shows that good features can be learned from weakly supervised data. Indeed, our models learn features that are nearly on par with those learned from an image collection with over a million manually defined labels, and achieve good results on a variety of datasets. (Obtaining state-of-the-art results requires averaging predictions over many crops and models, which is outside the scope of this paper.) Moreover, our results show that weakly supervised models can learn semantic structure from image-word co-occurrences. 

In addition, our results lead to three main recommendations for future work in learning models from weakly supervised data. First, our results suggest that the best-performing models on the Imagenet dataset are not optimal for weakly supervised learning. We surmise that current models have insufficient capacity for learning from the complex Flickr dataset. Second, multi-class logistic loss performs remarkably well in our experiments even though it is not tailored to multi-label settings. Presumably, our approximate multiclass loss works very well on large dictionaries because it shares properties with losses known to work well in that setting~\cite{mikolov13, usunier2009ranking, weston11}. Third, it is essential to sample data \emph{uniformly per class} to learn good visual features~\cite{akata14}. Uniform sampling per class ensures that frequent classes in the training data do not dominate the learned features, which makes them better suited for transfer learning.

In future work, we aim to combine our weakly supervised vision models with a language model such as word2vec \cite{mikolov13} to perform, for instance, visual question answering \cite{antol15,yu15}. We also intend to further investigate the ability of our models to learn visual hierarchies, such as the ``sports'' example in Section~\ref{Experiment 2}.

\subsection*{Acknowledgements}
We thank Ronan Collobert, Tomas Mikolov, Alexey Spiridinov, Rob Fergus, Florent Perronnin, L{\'e}on Bottou and the rest of the FAIR team for code support and helpful discussions.

\bibliographystyle{plain}
\bibliography{egbib}

\end{document}